\documentclass[conference]{IEEEtran}
\IEEEoverridecommandlockouts
\usepackage{cite}
\usepackage{amsmath,amssymb,amsfonts}
\usepackage{algorithmic}
\usepackage{graphicx}
\usepackage{textcomp}
\usepackage{xcolor}
\usepackage{amsmath}
\usepackage[nolist]{acronym}
\usepackage{siunitx}
\usepackage{bm}
\usepackage{mhchem}

\usepackage{graphicx}
\usepackage{amsmath} 
\usepackage{amssymb}  
\usepackage{mathrsfs}  
\usepackage{caption}
\usepackage{subcaption}

\usepackage{enumitem}
\setlist[itemize]{align=parleft,left=0pt..1em}

\def\BibTeX{{\rm B\kern-.05em{\sc i\kern-.025em b}\kern-.08em
    T\kern-.1667em\lower.7ex\hbox{E}\kern-.125emX}}

\usepackage{tikz}

\newcommand\copyrighttext{%
  \footnotesize \textcopyright \the\year{} IEEE. Personal use of this material is permitted. Permission from IEEE must be obtained for all other uses, in any current or future media, including reprinting/republishing this material for advertising or promotional purposes, creating new collective works, for resale or redistribution to servers or lists, or reuse of any copyrighted component of this work in other works.}

\newcommand\copyrightnotice{%
\begin{tikzpicture}[remember picture,overlay]
\node[anchor=south,yshift=10pt] at (current page.south) {\fbox{\parbox{\dimexpr0.75\textwidth-\fboxsep-\fboxrule\relax}{\copyrighttext}}};
\end{tikzpicture}%
}

\begin{document}

\title{Towards Drone-based Mapping of Volcanic Gases using Gas Tomography
}

\author{\IEEEauthorblockN{Marius Schaab\IEEEauthorrefmark{1}, Niklas Karbach\IEEEauthorrefmark{2}, Antonia Rabe\IEEEauthorrefmark{2}, Thomas Wiedemann\IEEEauthorrefmark{1}\IEEEauthorrefmark{3}, Patrick Hinsen\IEEEauthorrefmark{3},\\ Dmitriy Shutin\IEEEauthorrefmark{3},  Thorsten Hoffmann\IEEEauthorrefmark{2}, Achim J. Lilienthal\IEEEauthorrefmark{1}}
\thanks{
This work was facilitated by the German Research Foundation (DFG) Project SPP2433, which provided a platform for inter-institutional exchange and funding.
The authors also thank the Istituto Nazionale di Geofisica e Vulcanologia (INGV), especially Nicole Bobrowski, for their support. 
}
\IEEEauthorblockA{\IEEEauthorrefmark{1}School of Computation, Information and Technology, Technical University of Munich (TUM), Germany}
\IEEEauthorblockA{\IEEEauthorrefmark{2}Department of Chemistry, Johannes Gutenberg-University (JGU), Germany}
\IEEEauthorblockA{\IEEEauthorrefmark{3}Institute of Communications and Navigation, German Aerospace Center (DLR), Germany}\\

}
\IEEEaftertitletext{\vspace{-1.3cm}}
\maketitle
\copyrightnotice

\begin{abstract}
Volcanoes emit large amounts of \ce{CO2}, directly influencing human lives. Mapping volcanic gas emissions helps to forecast eruptions and understand the impact of volcanoes on climate and the environment. Drone-based gas sensing significantly reduces risks in volcanic monitoring but faces technical limitations when measuring gas, as rotor downwash disperses the gas plume before detection. Gas Tomography using remote gas sensing addresses this challenge. At the Salinelle dei Cappuccini mud volcanoes, we demonstrate that while drone-mounted in-situ sensors failed to detect \ce{CO2} emissions due to aerodynamic disturbance, open-path sensing successfully enabled remote gas distribution mapping. We present a novel model-based gas tomographic reconstruction approach that incorporates a Lagrangian model to compensate for wind-induced advection. The resulting gas distribution maps align with manually collected in-situ measurements, confirming that model-based gas tomography effectively overcomes downwash limitations and enables accurate mapping of volcanic emissions.
\end{abstract}


\section{Introduction}

\begin{figure}
    \centering
    \includegraphics[width=0.85\linewidth]{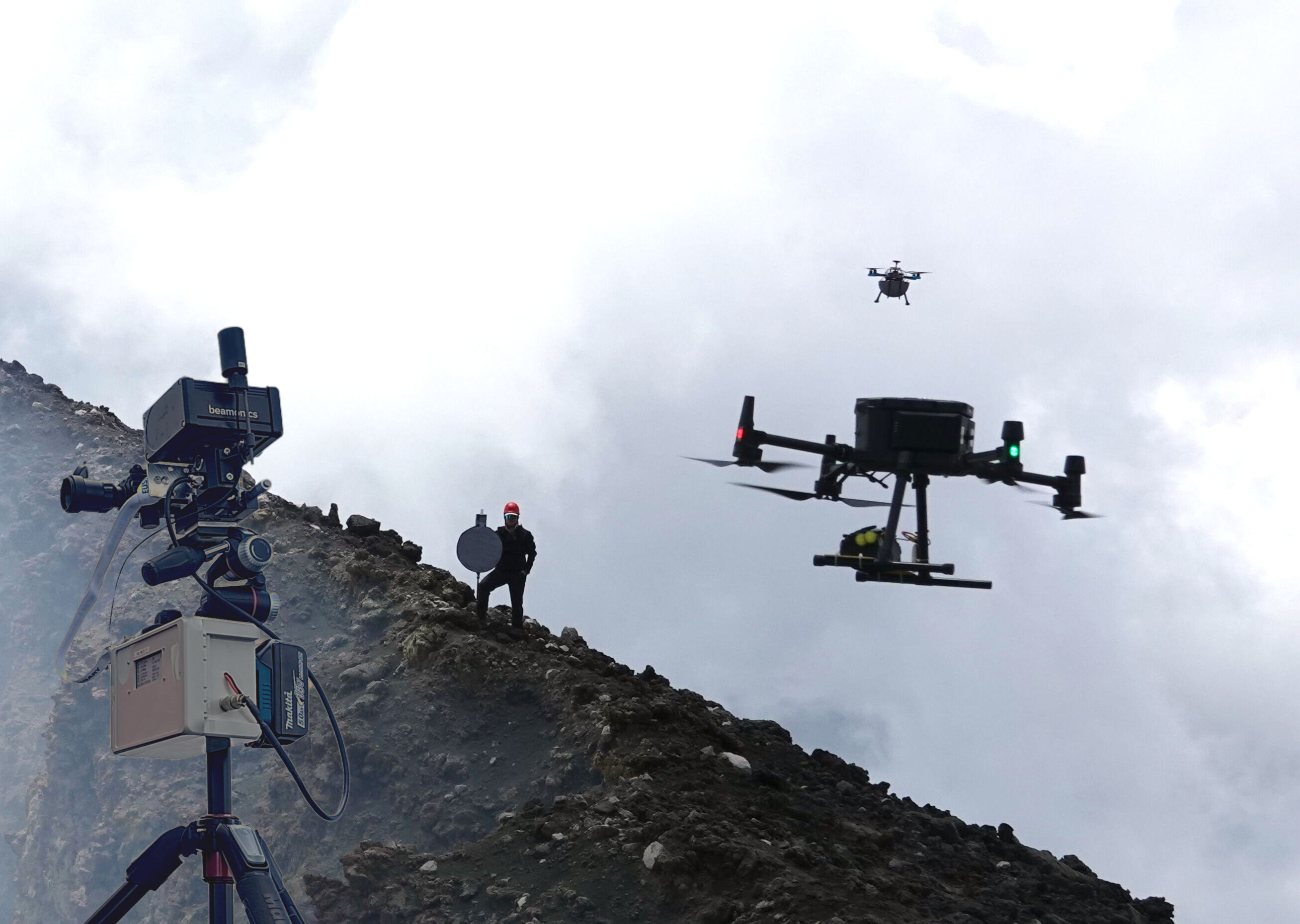}
    \caption{Measurements at Mount Etna. The TDLAS sensor (left) measures either against the handheld reflector in the background or the smaller drone with the spherical reflector. The larger drone carries an in-situ sensor.}
    \label{fig:setup}
\end{figure}

Measuring volcanic gases helps to predict eruptions and is crucial for modeling the impact of volcanoes on climate and the environment. 
Therefore, it is important to know the gas composition, identify potential degassing spots, and determine release rates to calculate the total amount of gas emitted into the environment. 
In-situ measurements at volcanoes involve some dangers. Researchers have to climb to the crater, exposing themselves to toxic gases, whereas drones enable faster measurements from a safe distance \cite{mainzer2025}. 
While a drone also allows for vertical profiling, its propulsion system introduces a significant disadvantage. The propeller downwash impairs the accuracy of gas measurements directly beneath the drone by diluting or dispersing the gas. 
To overcome the downwash problem, we introduce open-path sensors, more explicitly, Tunable Diode Laser Absorption Spectroscopy (TDLAS) sensors. A TDLAS sensor measures the concentration of one specific gas (e.g., \ce{CO2}) along a laser path between the sensor and a reflector. 
A single measurement expresses the integral of concentration along the laser beam in parts per million times meter (\unit{ppm*\metre}).
Mounting the reflector on a drone (Fig.~\ref{fig:setup}), combined with a fast sampling rate of up to \qty{100}{\hertz}, enables the coverage of large areas in a short time, while the drone itself does not need to get in contact with the gas, preventing aerodynamic disturbances.  
\begin{figure}
    \centering
    \includegraphics[width=0.8\linewidth]{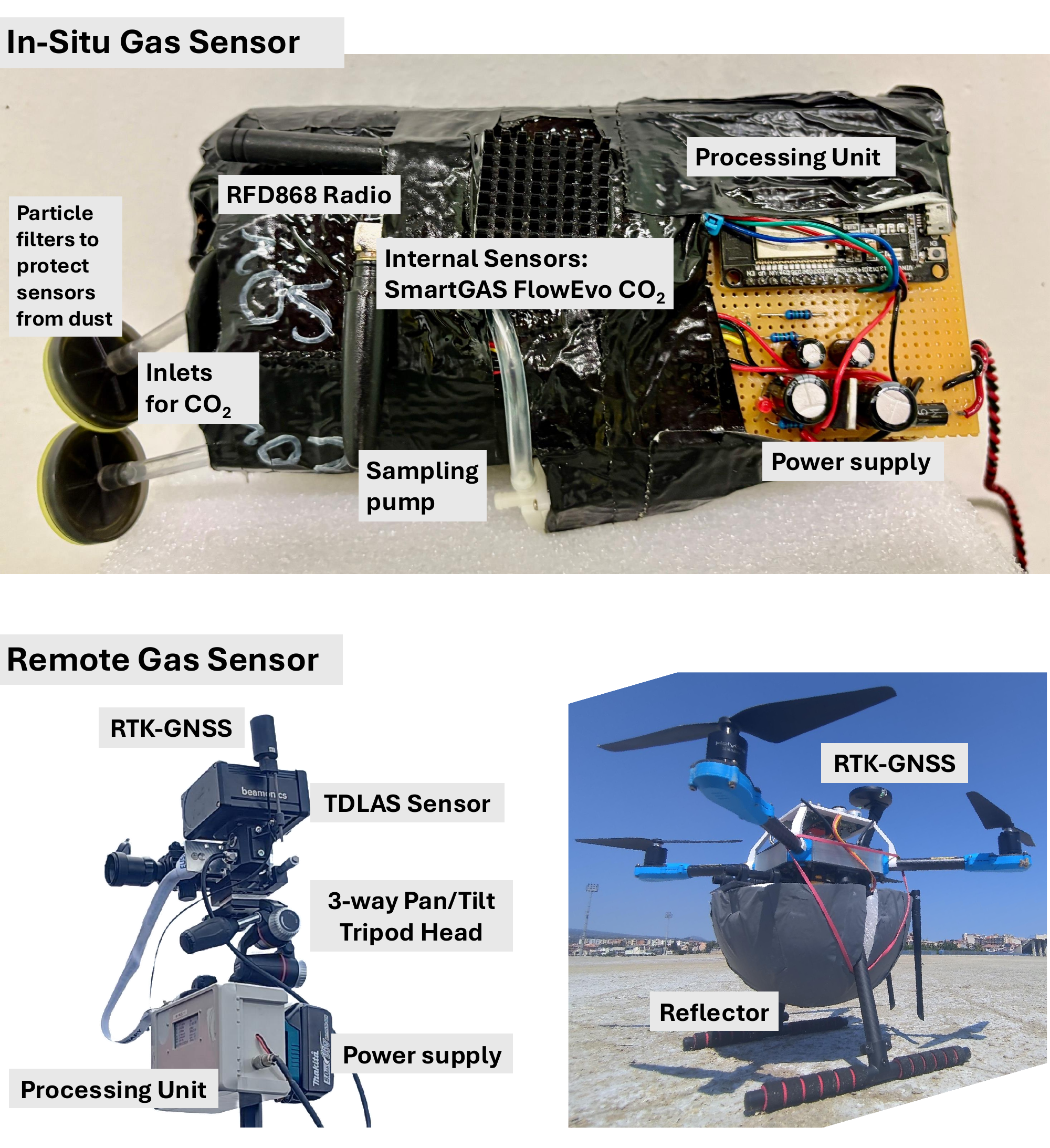}
    \caption{The top image depicts the in-situ sensor that can be mounted on a drone or carried by hand. The tape protects the sensor from the acidic volcanic environment. The bottom images show our two-component remote gas sensing system. The TDLAS sensor (left), model Beamonics BeamSight,  measures the gas concentration between itself and the reflector mounted on the drone (right).}
    \label{fig:sensors}
\end{figure}
This paper focuses on the challenges of drone-based open-path gas distribution mapping using model-based gas tomography. We focus on \ce{CO2} degassing close to the ground at the mud volcanoes Salinelle dei Cappuccini at the foot of Mount Etna in Paternò, Italy. The volcano serves as a sandbox for testing different measurement methods. 
The mud volcanoes emits mainly \ce{CO2} but also minor amounts of \ce{CH4}, \ce{H2}, and \ce{He}. The amount of \ce{CO2} varies between approximately \qtyrange{2e2}{2e4}{kg\per\day}, depending on the state of activity of the volcano\cite{ingv2019}. 
We conducted different experiments to evaluate the differences between drone-based and manual gas sampling, as well as in-situ and open-path sensing. We demonstrate the advantages of drone-based sampling, especially open-path sampling, over manual sampling.
Finally, we will introduce a simple method to mitigate the impact of wind on our gas tomography, which we then compare with the manual measurements obtained using an in-situ sensor.  

\section{Limitations of In-Situ Sensing}
\begin{figure}
    \centering
    \includegraphics[width=0.8\linewidth]{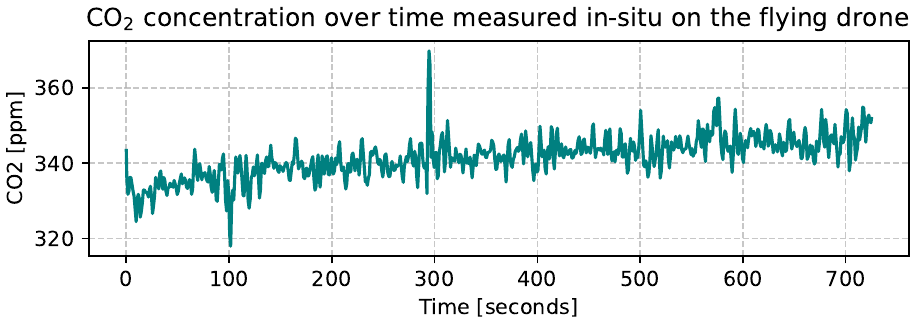}
    \caption{The figure shows the measurements of the in-situ sensor onboard the drone. The drone flew over the mud volcanoes and had only one \ce{CO2} detection at about \qty{300}{\second} after takeoff. }
    \label{fig:drone_no_measurements}
\end{figure}

Drones equipped with in-situ sensors offer the possibility of bringing measurement equipment into direct contact with airborne substances, such as gases or particles, and allow for measuring concentrations in the air along automated flight paths.
Using drones also significantly decreases the risk associated with measuring gases directly at the vent by diverting to better-located takeoff and landing sites, and makes terrain that is inaccessible by foot, accessible for measurement. Therefore, in a first experiment, a drone equipped with a SmartGAS FlowEvo \ce{CO2} sensor (\qtyrange{0}{2000}{ppm}) (Fig.~\ref{fig:sensors}) was used to approach individual degassing spots at the mud volcanoes.
This approach is commonly used to measure emissions from larger volcanoes, such as Vulcano (Aeolian Island, Italy) \cite{mainzer2022}. However, due to the significantly lower emission rates of the vents at the mud volcanoes, the downwash of the drone suppressed the gases from the sources below the drone, and prevented them from reaching the sensor, and in turn the sensor was not able to measure an increased \ce{CO2} level above the vents (Fig.~\ref{fig:drone_no_measurements}). 
Not only does the downwash impede measurements of small sources and low concentrations, but it also limits the precision of measurements of large-scale sources, where the plume is several tens of meters or even hundreds of meters in diameter.
Ambient air from above the drone is mixed with the sampled air from below the drone, interfering with gas distribution mapping and source localization.

In a second experiment, the sensor system was detached from the drone, and a manual walking tour was conducted around the area of the volcano, targeting visible vents on the ground. This walking tour was substantially harder than the drone flight, as parts of the surveyed area were flooded, had deep, mud-filled vents, or were not accessible at all. Due to the absence of the drone's downwash in this experiment, the sensor system could measure elevated \ce{CO2} values above the vents (Fig.~\ref{fig:handtomography}).

\section{Gas Tomography Measurements}
Remote gas sensing addresses downwash issues. The TDLAS sensor and the reflector remain outside of the plume, not affecting the natural dispersion of the gas plume \cite{neumann2018}. 
Additionally, remote gas sensing increases the chance of detecting gas: while an in-situ sensor only detects the gas concentration at a single point, a remote sensor integrates the gas concentration along its beam. This increased footprint of the sensor comes with the downside that we can no longer localize the gas along the beam. 
This can be mitigated by collecting multiple measurements from different angles, enabling us to perform gas tomography and reconstruct a gas distribution map in a given area\cite{trincavelli2012}.
By including model assumptions as described in \cite{schaab2025}, the gas distribution mapping becomes stable against different discretisations or resolutions of the map.

We collected two datasets using the TDLAS sensor to reconstruct 2D gas distribution maps close to the ground.
The sensor was mounted on a pan-tilt mechanism on a tripod, allowing us to manually aim at the reflector.
For the first dataset, we positioned the sensor at five different locations and aimed at a handheld reflector. Fig.~\ref{fig:measurements} shows the measurements. At each position, we sampled for approx. \qty{20}{\second} with \qty{5}{\hertz}. This set of measurements allows a two-dimensional tomography slice at a plane parallel to the ground at approx. \qty{1.5}{\meter} height. 
For the second dataset, we used a reflector mounted on the drone (Fig.~\ref{fig:sensors}). One operator manually piloted the drone, while another kept the laser beam aligned with the target. This system allows for rapid relocation of the reflector, even to different altitudes.
Both datasets took around \qty{1}{\hour} to acquire.

\setlength{\belowcaptionskip}{0pt}
\begin{figure*}
    \centering
    \begin{subfigure}[b]{0.3\linewidth}
    \centering
    \includegraphics[width=1\linewidth]{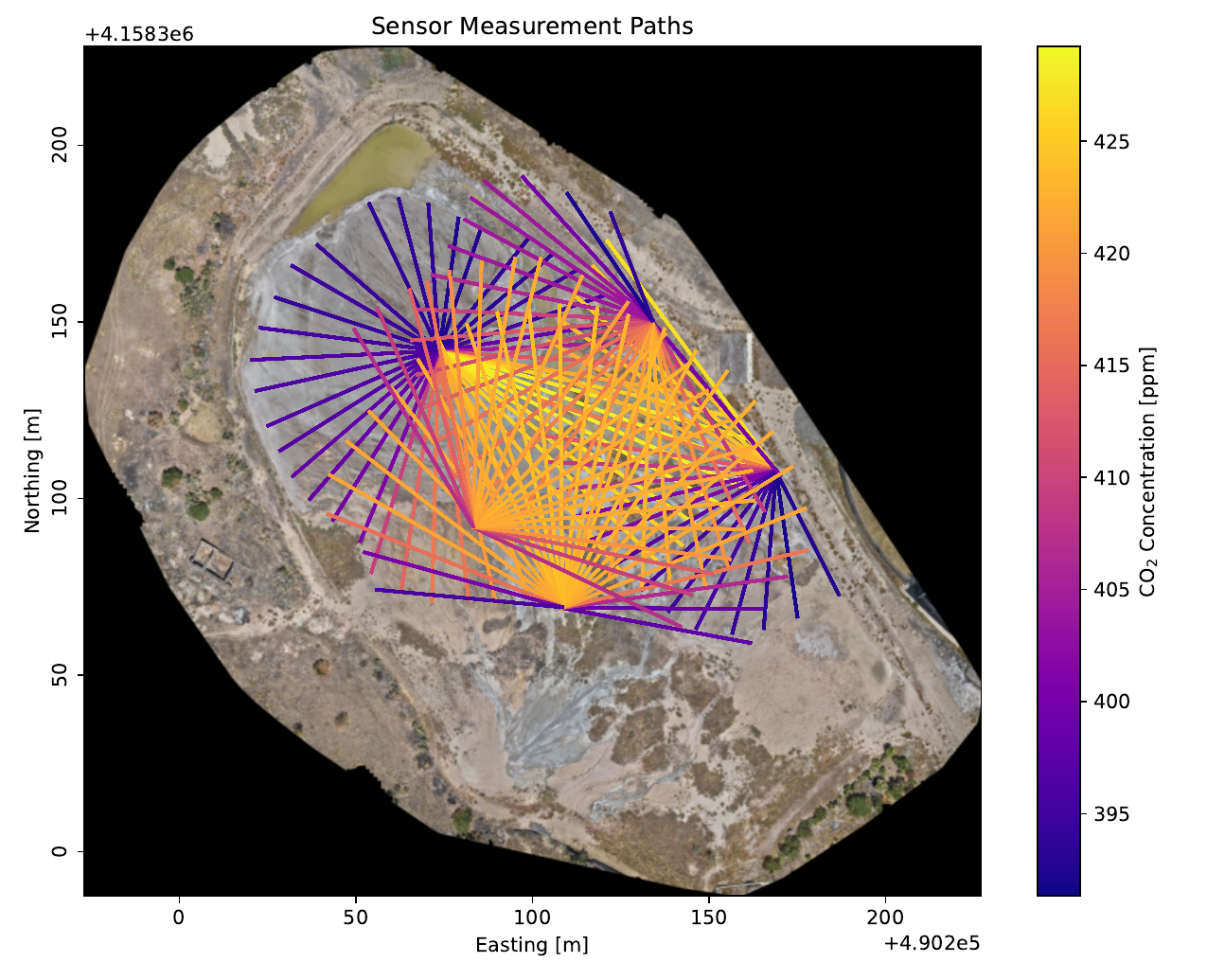}
    \caption{Handheld Reflector Measurements}
    \label{fig:measurements}
    \end{subfigure}
    \begin{subfigure}[b]{0.3\linewidth}
    \centering
    \includegraphics[width=1\linewidth]{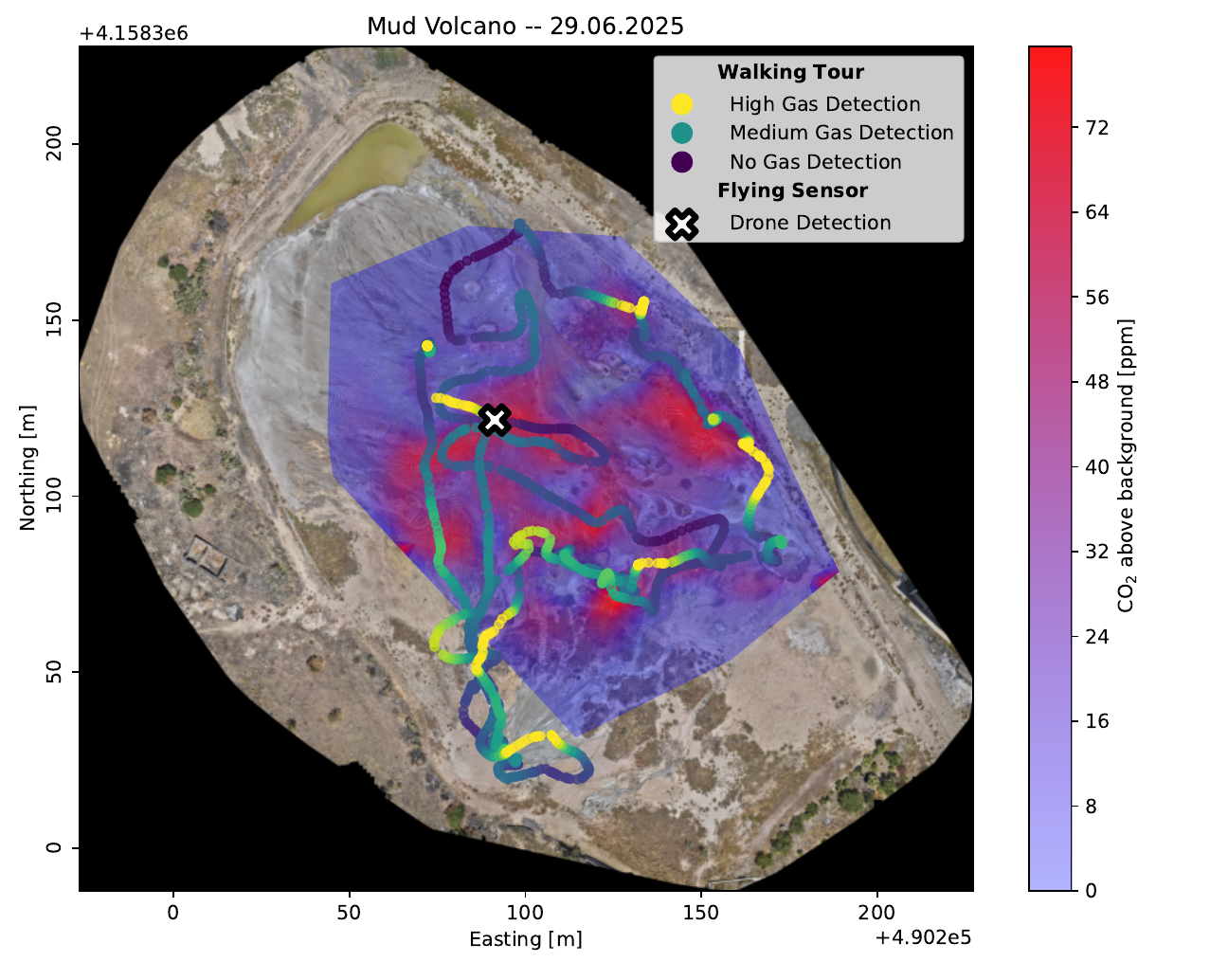}
    \caption{Handheld Reflector Reconstruction}
    \label{fig:handtomography}
    \end{subfigure}
    \begin{subfigure}[b]{0.3\linewidth}
    \centering
    \includegraphics[width=1\linewidth]{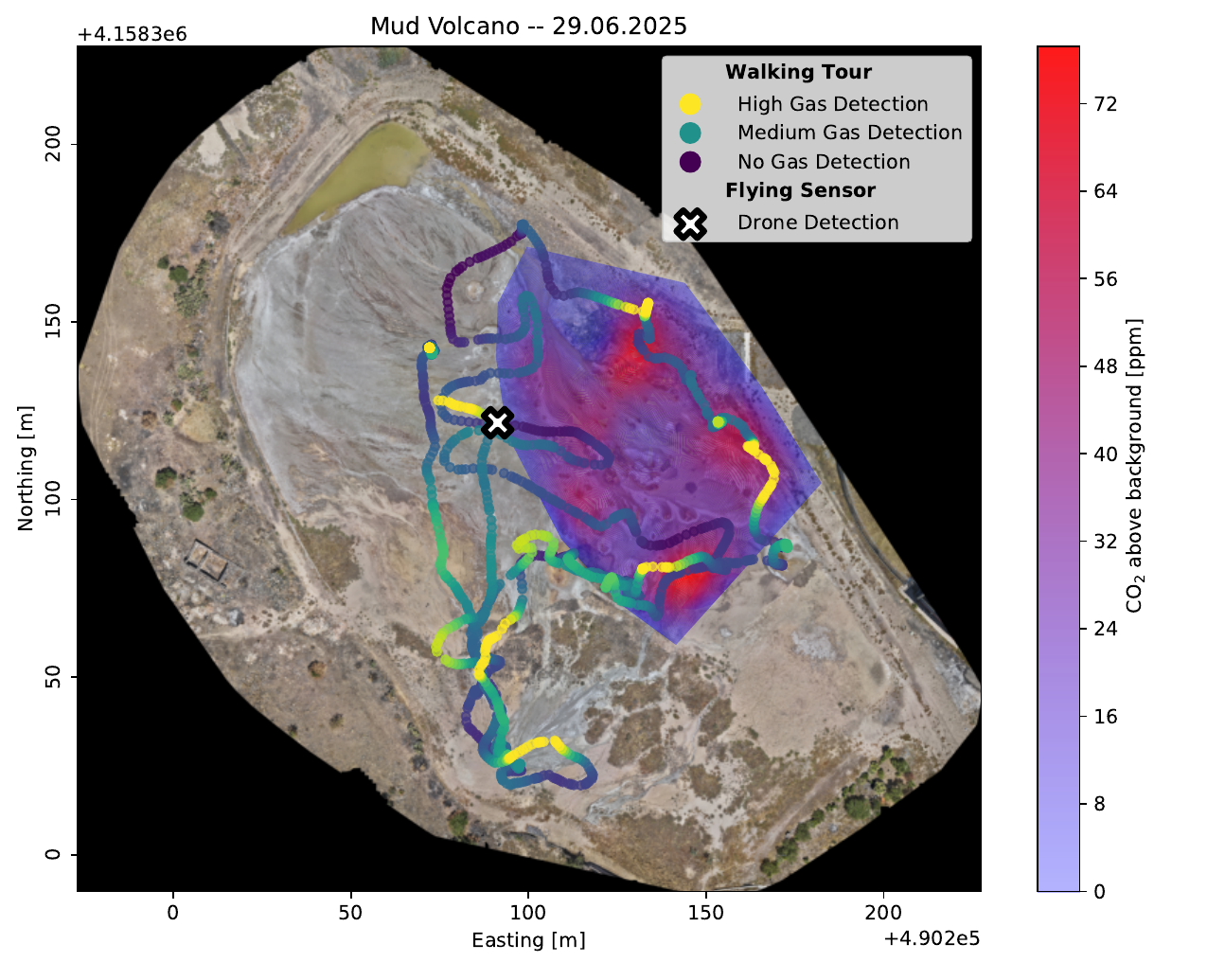}
    \caption{Drone Reflector Reconstruction}
    \label{fig:dronetomography}
    \end{subfigure}
    \setlength{\belowcaptionskip}{-10pt}
    \caption{Plot (a) shows the single open-path measurements using the handheld reflector. We used the measurements to calculate the gas distribution map in (b). Plot (c) shows the wind-compensated reconstruction from the data collected with the flying reflector. Both reconstruction plots include the gas detections of the in-situ sensors during the walking tour, as well as the single flying detection shown in Fig.~\ref{fig:drone_no_measurements}. The background map depicts a photogrammetry of the mud volcanoes.} 
    \label{fig:results}
\end{figure*}

\section{A simple Wind Compensation Method}
Gas Tomography with open-path sensors measures the gas distribution, not the source distribution.
However, the gas distribution measured very close to the ground can be assumed to show a high correlation with the actual source distribution and is therefore of particular interest.
Further above the ground, the current gas distribution primarily depends on the current wind field. 
During our measurements, the wind field constantly changed. 
In general, changing wind conditions lead to false reconstructions of the gas distribution using gas tomography. 
Therefore, we compensate for the changing wind conditions using a Lagrangian model, which describes the dispersion of gas based on the movement of molecules caused by advection due to wind and by diffusion.
We simplify by assuming no diffusion and a spatially constant wind.
On their way from the ground to our tomography measurement plane, the wind constantly displaces the emitted gas molecules, such that we measure these molecules at an offset location. Thus, we compensate by offsetting the sensor and reflector locations $\bm{x}^\text{TDLAS}_t,\ \bm{x}^\text{Reflector}_t$ with the wind vector~$\bm{w}_t$:
\setlength{\belowdisplayskip}{3pt} \setlength{\belowdisplayshortskip}{3pt}
\setlength{\abovedisplayskip}{3pt} \setlength{\abovedisplayshortskip}{0pt}
\begin{equation}
    \bm{x}^\text{new}_t = \bm{x}_t - \bm{w}_t \Delta t.
\end{equation}
We chose $\Delta t = \SI{3}{s}$ for our collected data at the mud volcanoes, meaning it took around \SI{3}{s} for the emitted gas to reach our measurement plane.
$\bm{w}_t$ is the averaged wind speed over timespan $[t-\SI{3}{s}, t]$.
We measured the wind using an anemometer located near our measurement area.

\section{Results}
Fig.~\ref{fig:results} shows the gas tomography reconstructions with the handheld and flying reflector. We computed the two-dimensional gas distribution map in Fig.~\ref{fig:handtomography} with our model-based gas tomography algorithm \cite{schaab2025} from the first dataset of the open-path TDLAS measurements (Fig.~\ref{fig:measurements}). Fig.~\ref{fig:dronetomography} depicts the gas distribution map calculated from the second dataset with the flying reflector. 
Fig.~\ref{fig:handtomography} and \ref{fig:dronetomography} include the gas detections of the hand-carried and drone-carried in-situ sensor. We observe that the locations of the gas detections generally match the gas distribution map. Note that our gas tomography reconstruction adjusts for wind, while in-situ sensor detections are plotted at their exact point of acquisition. Additionally, the gas tomography captures a time-averaged gas distribution map, whereas the in-situ sensor captures short snapshots at specific locations. Hence, gas tomography and in-situ sensing each measure different properties of the mud volcanoes, such that the significance of a direct comparison is limited. 
However, since there exists no ground-truth gas map of the mud volcano, a comparison between in-situ sensing and gas tomography serves as a reasonable option to evaluate the validity of the gas tomography.
Fig.~\ref{fig:results} additionally shows that both datasets have similar maximum values, and \ce{CO2} occurs at approximately the same locations. Human precision and the reflector size limit the maximum range of the open-path sensor; thus, the resulting reconstruction area of the drone-based dataset is smaller.

\section{Conclusion}

We evaluated drone-based gas sensing at the Salinelle dei Cappuccini mud volcanoes, demonstrating that propeller downwash limits in situ detection of gas sources beneath the drone. To overcome this limitation, we applied open-path TDLAS sensing combined with a Lagrangian wind compensation method.
This approach successfully reconstructed \ce{CO2} distribution maps that align with hand-collected in-situ measurements. Our wind compensation method introduces an additional tunable parameter, $\Delta t$. In the future, an optimization method that minimizes the error between TDLAS measurements and the reconstruction by varying $\Delta t$ could enhance this method. Additionally, a more advanced wind model would likely improve the results. 
Future measurement campaigns should also include a longer in-situ sensing period with more sensors to allow a significant comparison between gas tomography and an averaged map captured by in-situ sensors.

\begin{acronym}

\end{acronym}

\bibliographystyle{IEEEtran}

\bibliography{literature.bib}

\end{document}